\documentclass{article} 

\usepackage{amssymb}
\usepackage{amsmath}
\usepackage{amsthm}
\usepackage{titlesec}
\usepackage{subfigure} 

\usepackage{graphicx} 
\usepackage{algorithm}
\usepackage{algorithmic}

\title{Parametric Maxflows for Structured Sparse Learning with\\Convex Relaxations of Submodular Functions}

\author{
Yoshinobu Kawahara$^{\dagger}$,  and Yutaro Yamaguchi$^{\ast}$\\
$^{\dagger}$ The Institute of Scientific and Industrial Research, Osaka University,\\
$^{\ast}$  Department of Mathematical Informatics, The University of Tokyo,\\
\texttt{ykawahara@sanken.osaka-u.ac.jp, yutaro$\_$yamaguchi@mist.i.u-tokyo.ac.jp,}}
\date{}

\titlespacing{\paragraph}{%
  0pt}{
  0.3\baselineskip}{
  1em}%

\newtheorem{theorem}{Theorem}
\newtheorem{lemma}[theorem]{Lemma}
\newtheorem{proposition}[theorem]{Proposition}
\newtheorem{corollary}[theorem]{Corollary}
\newtheorem{definition}[theorem]{Definition}

\DeclareMathOperator*{\argmin}{argmin}

\DeclareMathOperator*{\prox}{prox}
\DeclareMathOperator*{\supp}{supp}

\begin{document}

\maketitle

\begin{abstract}
The proximal problem for structured penalties obtained via convex relaxations of submodular functions is known to be equivalent to minimizing separable convex functions over the corresponding submodular polyhedra. In this paper, we reveal a comprehensive class of structured penalties for which penalties this problem can be solved via an efficiently solvable class of parametric maxflow optimization. We then show that the parametric maxflow algorithm proposed by Gallo~et~al.~\cite{GGT89} and its variants, which runs, in the worst-case, at the cost of only a constant factor of a single computation of the corresponding maxflow optimization, can be adapted to solve the proximal problems for those penalties. Several existing structured penalties satisfy these conditions; thus, regularized learning with these penalties is solvable quickly using the parametric maxflow algorithm. We also investigate the empirical runtime performance of the proposed framework. 
\end{abstract}

\section{Introduction}
\label{sec:intro}

Learning with structural information in data has been a primary interest in machine learning.~Regularization with structured sparsity-inducing penalties, such as group Lasso \cite{YL06,JOV09} and (generalized) fused Lasso \cite{TSR+05,TT11}, has been shown to achieve high predictive performance and solutions that are easier to interpret, and has been successfully applied to a broad range of applications, including bioinfomatics \cite{MSH07,LL08,JOV09,KX12,SCL+13}, computer vision \cite{XKWG14,MBP14,SYT+15}, natural language processing \cite{ESX11,YS14} and signal processing \cite{SD11,TMN15}.

Recently, it has been revealed that many of the existing structured sparsity-inducing penalties can be interpreted as convex relaxations of submodular functions \cite{Bac10,OB12}. Based on this result, the calculation of the proximal operators for such penalties is known to be reduced to the minimization of separable convex functions over the corresponding submodular polyhedra, which can be solved via the iteration of submodular minimization. However, minimizing a submodular function is not effectively scalable (due to its generality); thus, an unavoidable next step is to clarify when the problem is solvable as a special case that can be calculated faster, especially cases that are solvable as an efficiently solvable class of network flow optimization. Several specific problems are known to be solvable via such network flow optimization. For example, a class of the total variation, which is equivalent to generalized fused Lasso (GFL), is known to be solved via parametric maxflows \cite{CD09,GY09}. Mairal~et~al.~(2011)~\cite{MJOB11} and Mairal~\&~Yu~(2013)~\cite{MY13} proposed parametric maxflow algorithms for $l_{1}/l_{\infty}$-regularization and the path-coding, respectively. In addition, Takeuchi~et~al.~(2015)~\cite{TKI15} recently proposed a generalization of GFL to a hyper-graph case, which they call higher-order fused Lasso, with a parametric maxflow algorithm.

In this paper, we first develop sufficient conditions for estimating whether a submodular function corresponding to a given structured penalty is graph-representable, i.e., realizable as a projection of a graph-cut function with auxiliary nodes. Several existing structured penalties from submodular functions, such as (overlapping) grouped penalty and (generalized) fused penalty, satisfy these conditions. Then, we show that the parametric maxflow algorithm proposed by Gallo~et~al.~\cite{GGT89} and its variants (hereafter, we call those the GGT-type algorithms) is applicable to calculate the proximal problems for penalties obtained via convex relaxation of such submodular functions, which runs at the cost of only a constant factor in the worst-case time bound of the corresponding maxflow optimization. Also, we empirically investigate the comparative performance of the proposed framework against existing algorithms.

Thus, the main contribution of this work is two-fold:\@ (i)~we develop sufficient conditions (with concrete ways of constructing the corresponding networks) for the class of structured penalties that can be solved via a parametric maxflow algorithm and (ii) we show that an efficient parametric flow algorithm can be applied to the proximal problem for such penalties. Note that the first one is closely related to the class of energy minimization problems that can be solved with the so-called graph-cut algorithm, which has been discussed actively in computer vision \cite{KZ04,KLT09}. Similar discussions are found in the context of realization of a submodular function as a cut function in combinatorial optimization \cite{BM85,Meg74,FP01}. Our current work would give a relation to such discussions to structured regularized learning. And as for the second one, our proposed formulation gives an unified view of the class of structured regularization that can be solved as a parametric maxflow problem, which generalizes, extends or connects several existing works that have been separately discussed to date, such as \cite{CD09,GY09,MJOB11,OB12,MY13,Bac13,XKWG14}, without increasing the essential theoretical run-time bound.

The remainder of this paper is organized as follows. We first define notations and describe preliminaries in Section~\ref{sec:notation}. Then, in Section~\ref{sec:relax}, we give a brief review of structured penalties obtained as convex relaxations of submodular functions. In Section~\ref{sec:network}, we describe the sufficient condition for estimating whether the proximal problem for a given penalty is solvable via network flow optimization. In Section~\ref{sec:paraflow}, we develop the parametric flow algorithm to proximal problems for penalties satisfying this condition. In Section~\ref{sec:related}, we describe related work.
Finally, we show runtime comparisons for calculating the proximal problem for the penalties by the proposed and existing algorithms in Section~\ref{sec:result}, and conclude the paper in Section~\ref{sec:concl}. All proofs are given in Appendix~\ref{app:proofs}.

\section{Notations and Preliminaries}
\label{sec:notation}

In this section, we introduce notations used in this paper, and give brief reviews on submodular functions in Section~\ref{ssec:submo} and network flow optimization in Section~\ref{ssec:flow}.

\subsection{Submodular Functions}
\label{ssec:submo}

Let $d$ be a positive integer and $V := \{1,2,\dots,d\}$. We denote the complement of $A$ by $\overline{A}$ for $A\subseteq V$, i.e., $\overline{A}=V\setminus A$.
For a real vector $\boldsymbol{w}=(w_i)_{i\in V}\in\mathbb{R}^V$ and a subset $A\subseteq V$, define $\boldsymbol{w}(A):=\sum_{i\in A}w_i$.
A set function $F\colon 2^V \to \mathbb{R}$ is called {\em submodular} if
\begin{equation*}
F(A)+F(B)\geq F(A\cap B)+F(A\cup B)
\end{equation*}
for any $A,B\subseteq V$ \cite{Edm70,Fuj05}.

We denote by $\widehat{F}$ the {\em Lov\'{a}sz extension} of a set function $F$ with $F(\emptyset)=0$, i.e., $\widehat{F} \colon \mathbb{R}^V \to \mathbb{R}$ is a continuous function defined as, for each $\boldsymbol{w} \in \mathbb{R}^V$,
\begin{equation*}
\widehat{F}(\boldsymbol{w}) := \sum_{i=1}^d w_{j_i}\bigl(F(\{j_1,\ldots,j_i\})-F(\{j_1,\ldots,j_{i-1}\})\bigr),
\end{equation*}
where $j_1, j_2, \dots, j_d \in V$ are the distinct indices corresponding to a permutation that arranges the entries of $\boldsymbol{w}$ in nonincreasing order, i.e., $w_{j_1}\geq w_{j_2} \geq\cdots\geq w_{j_d}$ \cite{Lov83}.

For a submodular function $F$ with $F(\emptyset)=0$, {\em the submodular polyhedron} $P(F)\subseteq\mathbb{R}^V$ and {\em the base polyhedron} $B(F)\subseteq\mathbb{R}^V$ are respectively defined as
\begin{align*}
P(F) :&= \{\,\boldsymbol{x}\in\mathbb{R}^V\mid\boldsymbol{x}(A)\leq F(A) ~(\forall A\subseteq V)\,\}
~~\text{and}~~ \\
B(F) :&= \{\,\boldsymbol{x}\in P(F)\mid \boldsymbol{x}(V)=F(V)\,\}.
\end{align*}
We define $P_+(F):=\mathbb{R}^V_{+}\cap P(F)$.

For an integer $i$ with $0\leq i\leq d$, let $\binom V i$ denote the set of $i$-element subsets of $V$. For any set function $F$, there uniquely exist functions $F^{(i)} \colon \binom V i \to \mathbb{R}$ $(i=0,1,\ldots,d)$ such that\vspace*{-2mm}
\begin{equation*}
F(A) = \sum_{i=0}^{|A|}\sum_{Y\in\binom A i} F^{(i)}(Y) ~~(A\subseteq V), \vspace*{-2mm}
\end{equation*}
where, for each $i=0,1,\ldots,d$,
\begin{equation*}
F^{(i)}(A) = \sum_{Y\subseteq A}(-1)^{|A-Y|}F(Y) ~~(A\in \textstyle\binom V i)
\end{equation*}
by the M\"{o}bius inversion formula (see, for example, \cite{Aig79}). A set function $F$ is said to be {\em of order $k$} for an integer $k$ with $0\leq k\leq d$ if $F^{(k)}\neq 0$ and $F^{(i)}=0$ ($k+1\leq i\leq d$).

\subsection{Flow Terminology}
\label{ssec:flow}

Suppose we are given a directed network $\mathcal{N}\,$$=(U,E)$ with a finite vertex set $U$ and an edge set $E\subseteq U \times U$, a distinguished {\em source vertex} $s\in U$, a distinguished {\em sink vertex} $t\in U$, and a nonnegative {\em capacity} $c(u,v)$ for each edge $(u,v)\in E$. Define $c(u, v) := 0$ for each pair $(u, v) \in (U \times U) \setminus E$. A {\em flow} $f$ on $\mathcal{N}$ is a real-valued function on vertex pairs satisfying the following three constraints:
\begin{align*}
f(u,v) &\leq c(u,v) &&\hspace*{-4mm}\text{for}\hspace*{-4mm}&& (u,v)\in U\times U &&\text{(capacity)}, \\
f(u,v) &= -f(v,u) &&\hspace*{-4mm}\text{for}\hspace*{-4mm}&& (u,v)\in U\times U &&\text{(antisymmetry)}, ~\text{and} \\
{\textstyle\sum_{u\in U}}f(u,v) &= 0 &&\hspace*{-4mm}\text{for}\hspace*{-4mm}&& v\in U\setminus\{s,t\} &&\text{(conservation)}.
\end{align*}
The value of flow $f$ is $\sum_{v\in U}f(v,t)$. A {\em maximum flow} is a flow of maximum value. For disjoint $A, B \subseteq V$, the {\em capacity} of pair $(A, B)$ is defined as $c(A,B):=\sum_{u\in A,v\in B}c(u,v)$. A {\em cut} $(C,\overline{C})$ is a vertex partition (i.e., $C\cup\overline{C}=U$, $C\cap\overline{C}=\emptyset$) such that $s\in C$ and $t\in\overline{C}$. A {\em minimum cut} is a cut of minimum capacity. The capacity constraint implies that for any flow $f$ and any cut $(C,\overline{C})$, we have $f(C,\overline{C})\leq c(C,\overline{C})$, which implies that the value of a maximum flow is at most the capacity of a minimum cut. The {\em max-flow min-cut theorem} of \cite{FF62} states that these two quantities are equal.

\section{Penalties via Convex Relaxation of Submodular Functions}
\label{sec:relax}

We briefly review structured penalties through convex relaxations of submodular functions, which cover several known structured sparsity-inducing penalties, in Subsection~\ref{ssec:penalty}, and then the existing optimization methods for those proximal problems in Subsection~\ref{ssec:prox}.

\subsection{Structured Penalties from Submodular Functions}
\label{ssec:penalty}

Structured penalties obtained via convex relaxations of submodular functions can be categorized into two types. Here, we review these respectively in Sections~\ref{sss:penal1} and \ref{sss:penal2}.

\subsubsection{Penalty via $\ell_p$-relaxation of Nondecreasing Submodular Function}
\label{sss:penal1}

The first type of the penalty from a submodular function is defined through convex relaxation with the $\ell_p$-norm \cite{Bac10,OB12,Bac13}. For this type, a submodular function $F$ is required to be non-decreasing. To define this penalty,~we first consider a function $h \colon \mathbb{R}^V \to \mathbb{R}$ that penalizes both supports and $l_p$-norm on the supports;
\begin{equation}
\label{eq:g}
h(\boldsymbol{w}) = \frac{1}{p}\|\boldsymbol{w}\|_p^p+\frac{1}{r}F(\supp(\boldsymbol{w})),
\end{equation}
where $1/p+1/r=1$. Note that when $p$ tends to infinity, function $g$ tends to $F(\supp(\boldsymbol{w}))$ restricted to the $l_\infty$-ball. The following is known for any $p\in(1,+\infty)$.
\begin{proposition}[\cite{OB12}]
\label{prop:norm}
Let $F$ be a non-decreaing function s.t.\@ $F(\{i\})>0$ for all $i\in V$. The tightest convex homogeneous lower-bound of $h(\boldsymbol{w})$ 
is a norm, denoted by $\tilde{\Omega}_{F,p}$, such that its dual norm equals to, for $\boldsymbol{s}\in\mathbb{R}^V$,
\begin{equation}
\label{eq:dual_norm}
\tilde{\Omega}_{F,p}^*(\boldsymbol{s}) = \sup_{A\subseteq V,A\neq\emptyset} \frac{\|\boldsymbol{s}_A\|_r}{F(A)^{1/r}}.
\end{equation}
\end{proposition}

Note that, if $F$ is submodular, then only stable inseparable sets may be kept in the definition of $\tilde{\Omega}_{F,p}^*$ in Eq.~\eqref{eq:dual_norm}. From the above definition, we obtain, for any $\boldsymbol{w}\in\mathbb{R}^V$,
\begin{align}
\label{eq:lp_sep}
\tilde{\Omega}_{F,p}(\boldsymbol{w})
&= \sup_{\boldsymbol{s}\in\mathbb{R}^V} \boldsymbol{w}^\top\boldsymbol{s} ~~\text{such that}~ \Omega_{F,p}^*(\boldsymbol{w})\leq 1 \notag\\
&= \sup_{\boldsymbol{s}\in\mathbb{R}^V} \boldsymbol{w}^\top\boldsymbol{s} ~~\text{such that}~ \forall A\subseteq V, \|\boldsymbol{s}_A\|_r^r\leq F(A) \notag\\
&= \sup_{\boldsymbol{t}\in P_+(F)}{\textstyle\sum_{i\in V}}t_i^{1/r}|w_i|,
\end{align}
where we change the variables as $t_i=s_i^r$. The first equality is obtained using the Fenchel duality. Consequently, the norm $\tilde{\Omega}_{F,p}$ is computed with a separable form over (the positive part of) the corresponding submodular polyhedron.

It is easy to check that, if we use $F(A)=|A|$, the $\tilde{\Omega}_{F,p}$ is equivalent to the $\ell_p$-regularization. And, if we use $F(A)=\sum_{g\in\mathcal{G}}\min\{|A\cap g|,1\}$ for a group of variables $\mathcal{G}$, then $\tilde{\Omega}_{f,p}$ is equivalent to the (possibly, overlapping) $\ell_1/\ell_\infty$ and non-overlapping $\ell_1/\ell_p$ group regularizations or provides group sparsity similar to the overlapping $\ell_1/\ell_p$ group regularization \cite{OB12,Bac13}.

\subsubsection{Penalty by the Lov\'{a}sz Extension of Submodular Function}
\label{sss:penal2}

The other type of penalty is defined as the Lov\'{a}sz extension, i.e., $\ell_\infty$-relaxation, of a submodular function~$F$ with $F(\emptyset)=F(V)=0$. This is known to make some of the components of $\boldsymbol{w}$ equal when used as a regularizer~\cite{Bac11}. A representative example of this type of penalty is {\em the generalized fused Lasso} (GFL), which is defined for a given undirected network $\mathcal{N}=(V,E)$ as
\begin{equation*}
\Omega_{\mathrm{fl}}(\boldsymbol{w}) = \sum_{(i,j)\in E} a_{ij}|w_i-w_j|,\end{equation*}
where $a_{ij}$ is the weight on each pair $(i,j)$. This penalty is known to be equivalent to the Lov\'{a}sz extension of a cut function on $\mathcal{N}$, i.e., $F(A) = \sum_{i\in A,j\in V\setminus A}a_{ij}$ \cite{Bac11,XKWG14}. This can be extended to a hypergraph $\mathcal{H}=(V,E)$ with non-negative weight $a_e$ for each hyperedge $e\in E$, where the Lov\'{a}sz extension of a hypergraph cut function $F(A)=\sum_{e\in E:e\cap A\neq\emptyset,e\cap\overline{A}\neq\emptyset}a_e$ gives the hypergraph regularization $\Omega_{\mathrm{hr}}(\boldsymbol{w})=\sum_{e\in E}a_e(\max_{i\in e}w_i-\min_{i\in e}w_j)^p$ \cite{HSJR13}.

From the definition, the Lov\'{a}sz extension of a submodular function with $F(\emptyset)=0$ can be represented as a greedy solution over the submodular polyhedron \cite{Lov83}, i.e.,
\begin{equation*}
\widehat{F}(\boldsymbol{w})  = \sup_{\boldsymbol{t}\in P_+(F)}\sum_{i\in V}t_i|w_i|.
\end{equation*}
which is in fact the equivalent form with Eq.~\eqref{eq:lp_sep} for $r=1$ (i.e., $p=\infty$).

\subsection{Proximal Problem for Submodular Penalties}
\label{ssec:prox}

The above penalties have a common form, for a (normalized) submodular function $F$,\begin{equation}
\label{eq:def_penal}
\Omega_{F,p}(\boldsymbol{w}):= \sup_{\boldsymbol{t}\in P_+(F)}\sum_{i\in V}t_i^{1/r}|w_i|,
\end{equation}
where $p\in(1,+\infty)$ and $1/p+1/r=1$. However, note that, if $F$ is not nondecreasing, then $\Omega_{F,p}(\boldsymbol{w})$ does not necessarily has the duality as described in Section~\ref{ssec:penalty}. When using the norm $\Omega_{F,p}$ as a regularizer, we solve the following problem for some (convex and smooth) loss $l \colon \mathbb{R}^V \to \mathbb{R}$ that corresponds to the respective learning task:
\begin{equation*}
\min_{\boldsymbol{w}\in\mathbb{R}^V}l(\boldsymbol{w})+\lambda\cdot\Omega_{F,p}(\boldsymbol{w}) ~~(\lambda>0).
\end{equation*}
Since the objective of this problem is the sum of smooth and non-smooth convex functions, a major option for its optimization is the proximal gradient method, such as FISTA (Fast Iterative Shrinkage-Thresholding Algorithm) \cite{BT09}. Thus, our necessary step is to compute iteratively the proximal operator
\begin{equation}
\label{eq:prox_lp}
{\textstyle\prox_{\lambda\Omega_{F,p}}}(\boldsymbol{z}) := \argmin_{\boldsymbol{w}\in\mathbb{R}^d}\frac{1}{2}\|\boldsymbol{z}-\boldsymbol{w}\|_2^2 + \lambda\cdot\Omega_{F,p}(\boldsymbol{w}),
\end{equation}
where $\boldsymbol{z}\,$$\in\,$$\mathbb{R}^V$. From the definition \eqref{eq:def_penal}, we can calculate $\prox_{\lambda\Omega_{F,p}}$ by solving\begin{align}
\min_{\boldsymbol{w}\in\mathbb{R}^V}\max_{\boldsymbol{t}\in P_+(F)}\frac{1}{2}\|\boldsymbol{w}-\boldsymbol{z}\|_2^2+\lambda\sum_{i\in V}t_i^{1/r}|w_i|
&= \max_{\boldsymbol{t}\in P_+(F)}\sum_{i\in V}\min_{w_i\in\mathbb{R}}\left\{\frac{1}{2}(w_i-z_i)^2+\lambda t_i^{1/r}|w_i|\right\} \notag\\
&= -\min_{\boldsymbol{t}\in P_{+}(F)}\sum_{i\in V}\psi_{i}(t_{i}),\label{eq:sepa_prob}
\end{align}
where $\psi_i(t_i)=-\min_{w_i\in\mathbb{R}}\{\frac{1}{2}(w_i-z_i)^{2}+\lambda t_i^{1/r}|w_i|\}$. Thus, solving the proximal problem equals minimizing a separable convex function over the submodular polyhedron.

Based on the above formulation, Obozinski~\&~Bach~(2012)~\cite{OB12} recently suggested a divide-and-conquer algorithm as an adaptation of the decomposition algorithm by Groenevelt~(1991)~\cite{Gro91} for penalties from general submodular functions (for the case of $p=2$). A more general version of this approach was also developed by Bach~(2013)~\cite{Bac13}. However, a straightforward implementation of this approach yields $O(d)$-time calculation of submodular minimization, which could be time-consuming especially in large problems.

We address this issue by considering it from the following two perspectives. First, in Section~\ref{sec:network}, we develop an explicit sufficient conditions for determining whether the proximal problem for a given penalty can be solved through maximum flow optimization rather than submodular minimization. Maximum flow optimization can be regarded as an efficiently-solvable special case of submodular minimization, and is known to be much faster than submodular minimization in general; thus, this could be useful to judge whether a given penalty can be dealt with in a scalable manner as a regularizer. The respective structured penalties from submodular functions mentioned above are in fact instances of this case. On that basis, in Section~\ref{sec:paraflow}, we develop a procedure for problem~\eqref{eq:prox_lp} that runs at the cost of only a constant factor in its worst-case time bound of the maxflow calculation rather than the $O(d)$-time calculation of the straightforward implementation. In other words, we discuss whether an efficient parametric maxflow algorithm is applicable to the current problem.

\section{Graph-Representable Penalties}
\label{sec:network}

In this section, we develop sufficient conditions for determining whether the proximal problem for a given structured penalty is solvable through an efficiently-solvable class of network flow optimization. We also describe a concrete procedure to construct the corresponding network.

\subsection{Graph-Representable Set Functions}
\label{ssec:representable_functions}

The currently-known best complexity of minimizing a general submodular function is $O(d^6+d^5\,\text{EO})$, where EO is the cost of evaluating a function value \cite{Orl09}. Although there exist practically faster algorithms, such as the minimum-norm-point algorithm \cite{FHI06} as well as faster algorithms for special cases (e.g., Queyranne's algorithm for symmetric submodular functions \cite{Que98}), their scalability would not be practically sufficient, especially if we must solve submodular minimization several times, which is the current case. In addition, it is well known that 
a cut function (which is almost equivalent to a second order submodular function \cite{FP01}) can be minimized much faster through calculation of maxflows over the corresponding network. Given a directed network $\mathcal{N}=(V,E)$ with nonnegative capacity $c(e)$ on each edge $e\in E$, a cut function $\kappa_{\mathcal{N}}\colon V \to \mathbb{R}$ is defined as
\begin{equation*}
\kappa_\mathcal{N}(A) := \sum\{\, c(e) \mid e\in\delta_\mathcal{N}^{\text{out}}(A) \,\}~ ~ (A \subseteq V),
\end{equation*}
where $\delta_\mathcal{N}^{\text{out}}(A)$ denotes the set of edges leaving $A$ in $\mathcal{N}$. If $\mathcal{N}$ consists of $d$ nodes and $m$ edges, the currently best runtime bound for the minimization is $O(md)$ \cite{Orl13}. Albeit it is a better run-time bound, the empirical complexity is often much better with practical fast algorithms, e.g., \cite{GT88,BVZ01}.

However, the expressive power of a cut function is limited. Therefore, in order to balance between expressiveness and computational simplicity, using a higher-order function that is represented as a cut function with auxiliary nodes is often helpful. Such a function is sometimes referred to as {\em graph-representable} \cite{JLB11},\footnote{This class of functions is closely related to the class of energy minimization problems that can be solved by the so-called graph-cut algorithm \cite{BVZ01,KZ04}. Related results are also found in the context of realization of a submodular function as a cut function in combinatorial optimization \cite{BM85,FP01}.} 
and defined as follows. Let $U=V\cup W\cup \{s,t\}$ for some finite set $W$ with $W \cap V = \emptyset$ and distinct elements $s,t\not\in V\cup W$, and let $\tilde{\mathcal{N}}=(U,\tilde{E})$ be a directed network with nonnegative capacity $c(e)$ on each edge $e\in\tilde{E}$. Then, define a set function $F \colon 2^V \to \mathbb{R}$ as
\begin{equation*}
F(A) := \min_{Y\subseteq W}\kappa_{\tilde{\mathcal{N}}}(\{s\}\cup A\cup Y) + C_F \quad (A \subseteq V),
\end{equation*}
where $C_F \in \mathbb{R}$ is an arbitrary constant, and such $F$ is said to be graph-representable. If $W$ is empty, this function coincides with a cut function. The submodularity of this function is derived from the classical result of Megiddo~(1974)~\cite{Meg74} on network flow problems with multiple terminals (see, for the proof, \cite{NK13}).

\begin{figure}[t]\hspace*{-7mm}
\begin{minipage}{.268\linewidth}
\subfigure[Network from Condition (i)]{
\includegraphics[width=1.15\linewidth]{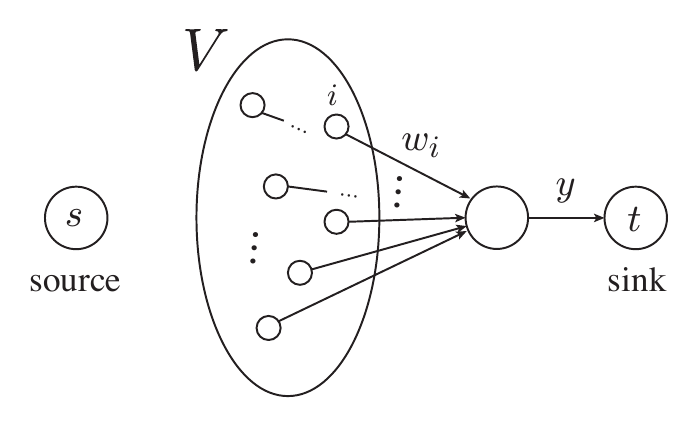}
\label{sfig:theorem1}}
\end{minipage}
\begin{minipage}{.73\linewidth}
\begin{minipage}{.47\linewidth}
\subfigure[Overview (Conditions (ii) and (iii))]{
\includegraphics[width=1.25\linewidth]{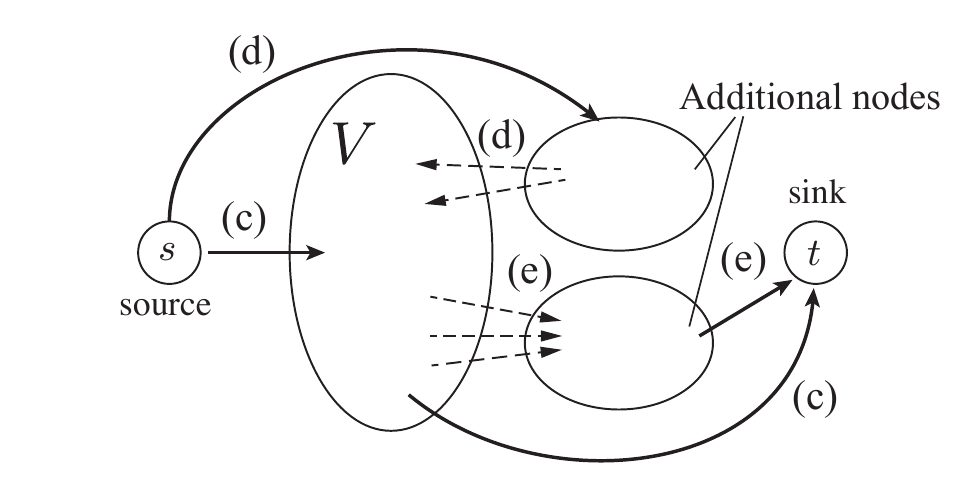}
\label{sfig:overview}}
\\\vspace*{4mm}\\
\subfigure[Linear term (each $v\in V$)]{
\label{sfig:lin}
\includegraphics[width=1.05\linewidth]{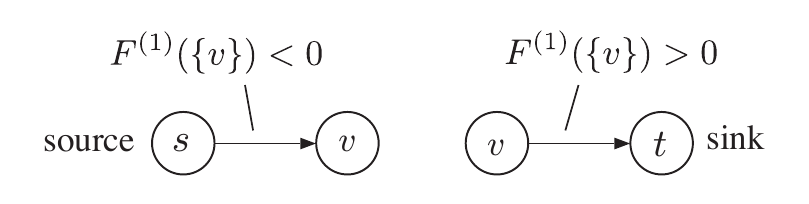}
\label{sfig:linear}}
\end{minipage}
\begin{minipage}{.528\linewidth}
\hspace*{13mm}
\subfigure[Second and negative third or higher order terms]{
\label{sfig:second}
\includegraphics[width=.65\linewidth]{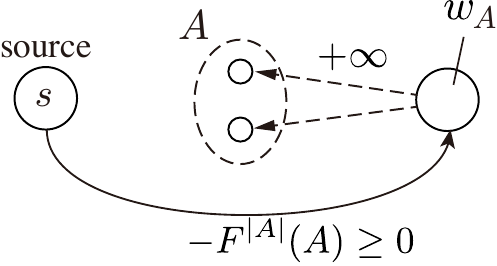}
\label{sfig:w3_pos}}
\begin{center}
\subfigure[Positive third order terms]{
\label{sfig:others}
\includegraphics[width=1.2\linewidth]{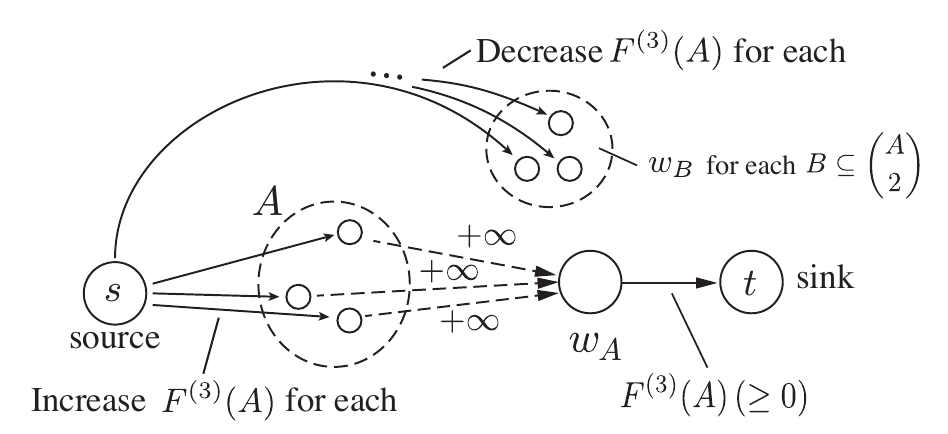}
\label{sfig:w3_neg}}
\end{center}\end{minipage}
\end{minipage}
\caption{Construction of network $\tilde{\mathcal{N}}$ for a graph-representable function in Theorem~\ref{th:sufficient} (Left(a):~Condition (i) and Right (b)-(e):~Conditions (ii) and (iii)).}
\label{fig:graph}
\end{figure}

\subsection{Sufficient Conditions and Network Construction}
\label{ssec:cond}

As described in Section~\ref{sec:paraflow}, if the corresponding set function $F$ for norm $\Omega_{F,p}$ is graph-representable, then its proximal problem~\eqref{eq:prox_lp} can be efficiently solved through a parametric maxflow computation. Hereafter, we refer to such a penalty as a {\em graph-representable penalty}, which is defined as follows.
\begin{definition}[Graph-representable penalty]
A penalty defined in Proposition~\ref{prop:norm} is said to be {\em graph-representable} if the set function $F$ on supports is graph-representable.
\end{definition}
Here, we present three types of sufficient conditions for a penalty $\Omega_{F,p}$ from a given submodular function $F$ (as described in Section~\ref{ssec:penalty}) to be graph-representable by constructing networks representing $F$.
The first one is mentioned as ``truncations,'' where a function $F$ is graph-representable by just one additional node (see Figure~\ref{sfig:theorem1} and also refer \cite{JLB11} or \cite{NK13}).
The second one is closely related to \cite{BM85}, and the third one is derived from \cite{FP01},
for which we describe concrete procedures to construct networks (see Figure~\ref{fig:graph} for the construction).

\begin{theorem}
\label{th:sufficient}
A set function $F$ with one of the following conditions is graph-representable.
\begin{itemize}
  \setlength{\itemsep}{-.5mm}
\item[{\rm (i)}]
  $F(A) = \min\{\boldsymbol{w}(A), y\}$ for some $\boldsymbol{w} \in \mathbb{R}_+^V$ and $y \in \mathbb{R}_+$.
\item[{\rm (ii)}]
  $F$ is submodular and of order at most three, i.e., $F^{(i)} = 0\,$ for $i = 4, 5, \ldots, d$.
\item[{\rm (iii)}]
  $F$ has no positive term of order at least two, i.e., $F^{(i)} \leq 0\,$ for $i = 2, 3, \ldots, d$.
\end{itemize}
\end{theorem}

\noindent{\em Remark.}~
It should be noted that
the sum of graph-representable submodular functions is also graph-representable
by considering the union of the corresponding networks.

\subsection{Examples}
A submodular function $F(A) = \sum_{g\in\mathcal{G}}\min\{|A\cap g|,1\}$, which gives the grouped-type regularization, is graph-representable since each term $\min\{|A\cap g|,1\}=\min\{\boldsymbol{e}_g(A),1\}$ is guaranteed to be so from Condition (i). The cost for constructing the corresponding network for this function is $O(|\mathcal{G}|)$ and the number of the additional nodes is $|\mathcal{G}|$.

Condition (ii) is a generalization of the condition that a cut function $F(A)=\sum_{i\in A,j\in V\setminus A}a_{ij}$ for a network $(V,E)$ can be solved with maximum flows, i.e.,  positive weights $a_{ij}$ for all $i,j\in E$.

Besides, a hypergraph cut function $F(A)=\sum_{e\in E:e\cap A\neq\emptyset,e\cap\overline{A}\neq\emptyset}a_e$ is also confirmed as graph-representable as follows. For each hyperedge $e \in E$, define $F_{e, 1}(A) := a_e\cdot\min\{|A \cap e|, 1\}$, $F_{e, 2}(A) := -a_e$ if $e \subseteq A$, and $F_{e, 2}(A) := 0$ otherwise. Then, $F_{e, 2}^{(|e|)}(e) = -a_e$ and $F_{e, 2}^{(|A|)}(A) = 0$ for $A \neq e$. Hence, $F_{e, 1}$ and $F_{e, 2}$ satisfy Conditions (i) and (iii), respectively, and it is easy to see $F = \sum_{e \in E} (F_{e, 1} + F_{e, 2})$. The network construction requires $O(\|E\|)$ time, where $\|E\| = \sum_{e \in E}|e|$.

\section{Parametric Maxflows for $\prox_{\Omega_{F,p}}(\boldsymbol{z})$}
\label{sec:paraflow}

We describe how the proximal problem~\eqref{eq:prox_lp} for a network representable penalty is solvable with  an adaptation of the GGT-type algorithms. We first derive a parametric formulation of this problem in Subsection~\ref{ssec:para_form}, and then develop the algorithm in Subsection~\ref{ssec:algo}.

\subsection{Parametric Formulation}
\label{ssec:para_form}

We address a parametric formulation of problem~\eqref{eq:sepa_prob}. To this end, we first consider
\begin{equation}
\label{eq:over_B}
\min_{\boldsymbol{\tau}\in B_+(F)}{\textstyle\sum_{i\in V}}\psi_i(\tau_i).\vspace*{-1mm}
\end{equation}
Note that the above optimization is over $B_+(F)$ in place of $P_{+}(F)$. In the following parts of this section, we suppose that $F$ is non-decreasing (thus, $B_+(F)$ coincides with $B(F)$). Although this does not necessarily hold for our case, we can show the following:
\begin{lemma}
\label{lem:non_dec}
Let $\boldsymbol{b}\in\mathbb{R}^V$ and $F$ be submodular, and set $\beta:=\sup_{i\in V}\{0,F(V\setminus\{i\})-F(V)\}/b_i$. Then, $F+\beta\boldsymbol{b}$ is a nondecreasing submodular function. Also, $\boldsymbol{\tau}^*$ is optimal to problem \eqref{eq:over_B} for $F$ if and only if $\boldsymbol{\tau}^*+\beta\boldsymbol{b}$ is optimal to problem \eqref{eq:over_B} for $F+\beta\boldsymbol{b}$.
\end{lemma}
Thus, for $F$ that is not non-decreasing, we can apply the algorithm developed below and recover an optimal solution to the original problem by transforming it to a non-decreasing one as in this lemma.

First, we define an interval $J\in\mathbb{R}$ as
\begin{equation*}
J := \bigcap_{i\in V}\{\, \psi_i'(\tau_i) \mid \tau_i\in(\text{dom}\, \psi_i\cap \mathbb{R}_+) \,\} ~(=(-\infty,0]).
\end{equation*}
Let $\boldsymbol{\tau}^*$ be an optimal solution to problem~\eqref{eq:over_B}. Denote the distinct values of $\psi_i'(\tau_i^*)$ by $\xi_{1}^{*}<\cdots<\xi_{k}^{*}$, and let $\xi_{0}^{*}:=-\infty$ and $\xi_{k+1}^{*}:=+\infty$. Let $A_{j}^{*}:=\{\, i \in V \mid \psi_{i}(\tau_{i}^{*})\leq \xi_{j}^{*} \,\}$ for $j=0,1,\ldots,k+1$. Also, let\begin{equation*}
F_{\alpha}(A):=F(A)-{\textstyle\sum_{i\in A}}\phi_i(\alpha) ~~(\alpha\in J),
\end{equation*}
where $\phi_i(\alpha)=\psi_i'^{-1}(\alpha) \,(\alpha\in J\setminus\{0\})~\text{or}~(|z_i|/\lambda)^r \,(\alpha=0$), and $\bullet^{-1}$ means an inverse function.
\begin{lemma}
Let $\alpha\in J$. If $\xi_{j}^{*}<\alpha<\xi_{j+1}^{*}$, $A_{j}^{*}$ is a minimizer of $F_{\alpha}$. If $\alpha=\xi_{j}^{*}$, $A_{j-1}^{*}$ is a minimal minimizer and $A_{j}^{*}$ is a maximal minimizer of $F_{\alpha}$.\end{lemma}
This is obtained, in Lemma~4 of \cite{NA12}, by replacing the assumption on the strict convexity of $\psi_{i}'$ with the monotonisity of the function in the region under consideration. As discussed in \cite{NA12}, this lemma implies that problem~\eqref{eq:sepa_prob} can be reduced to the following parametric problem:
\begin{equation}
\label{eq:para_prob}
\min_{A\subseteq V}F_{\alpha}(A) ~~\text{for all}~ \alpha\in J.
\end{equation}
That is, once we have the chain of solutions $A_0^*\subset\cdots\subset A_{k+1}^*$ to problem~\eqref{eq:para_prob} for all $\alpha\in J$, we can obtain an optimal solution to problem~\eqref{eq:over_B} as for $j=0,\ldots,k$
\begin{equation}
\label{eq:A2tau}
\tau_i^* = \phi_i(\alpha_{j+1}^*) ~(i\in A_{j+1}^*\setminus A_j^*)
~~\text{with}~\alpha_{j+1}^*
~~\text{s.t.}~F(A_{j+1}^*)-F(A_j^*)={\textstyle \sum_{i\in A_{j+1}^*\setminus A_j^*}}\phi_i(\alpha).
\end{equation}
The key here is that, if function $F$ is graph-representable, problem~\eqref{eq:para_prob} can be solved as a parametric minimum-cut  (equivalently, a parametric maxflow) problem on $\tilde{\mathcal{N}}$, where $c(s,v)$ for $v\in V$ are functions of $\alpha$ (since $\phi_i(\alpha)\geq 0$ for $\alpha\in J$), as will be stated in the next subsection.

Once we have a solution $\boldsymbol{\tau}^*$ to problem~\eqref{eq:over_B}, we can then obtain a solution to problem~\eqref{eq:prox_lp} as follows.
\begin{corollary}
\label{cor:w_star}
If $\boldsymbol{\tau}^*$ be an optimal solution to problem~\eqref{eq:over_B}, then the one to problem~\eqref{eq:prox_lp} is given by
\begin{equation*}
w_i^* = \begin{cases}
z_i-\text{\rm sign}(z_i)\lambda (\max(\tau_i^*,0))^{1/r} \hspace*{-2mm}& \text{if}~~ 0\leq \tau_i^*\leq (|z_i|/\lambda)^r, \\
0 & \text{otherwise}.
\end{cases}
\end{equation*}
\end{corollary}

\subsection{Algorithm Description}
\label{ssec:algo}

\begin{algorithm}[t]
\begin{algorithmic}[1]
\renewcommand{\algorithmicrequire}{\textbf{Input:}} \REQUIRE $\boldsymbol{z}\,$$\in\,$$\mathbb{R}^d$, $\tilde{\mathcal{N}}=(U,E)$.~ {\bf Output:} $\boldsymbol{w}^{*}\,$$=\,$$\prox_{\lambda\Omega_{F,p}}(\boldsymbol{z})$.
\STATE Compute $\alpha_0$ as in Eq.~\eqref{eq:alpha0} and set $\alpha_{k+1}\leftarrow 0$. Compute maximum flows $f_0$ and $f_{k+1}$, and minimum cuts $(C_0,\overline{C}_0)$ and $(C_{k+1},\overline{C}_{k+1})$ for $\alpha_0$ and $\alpha_{k+1}$ such that $|C_0|$ and $|C_{k+1}|$ are maximum and minimum by applying the preflow algorithm to $\tilde{\mathcal{N}}$, respectively. Form $\mathcal{N}'$ from $\tilde{\mathcal{N}}$ by shrinking the nodes in $C_0$ and in $\overline{C}_{k+1}$ to single nodes respectively, eliminating loops, and combining multiple arcs by adding their capacities.
\STATE If $\mathcal{N}'$ has at least three vertices, let $f_0'$, $f_{k+1}'$ be respectively the flows in $\mathcal{N}'$ corresponding to $f_0$, $f_{l+1}$. Then, perform {\em Slice}$(\mathcal{N}',\alpha_0,\alpha_{k+1},f'_0,f'_{k+1},C_0,C_{k+1})$.
\STATE Compute $\boldsymbol{w}^*$ as in Corollary~\ref{cor:w_star} and return $\boldsymbol{w}^*$.
\end{algorithmic}
\vspace*{1mm}{\bf Procedure} {\em Slice}$(\mathcal{N},\alpha_l,\alpha_{u},f_l,f_{u},A_l,A_u)$\hspace*{0mm}
\begin{algorithmic}[1]
\STATE Find $\tilde{\alpha}$ such that $c_{\tilde{\alpha}}(\{s\},U\setminus\{s\})=c(U\setminus\{t\},\{t\})$ (cf.~Lemma~\ref{lem:param}).
\STATE Run the preflow algorithm for $\tilde{\alpha}$ on $\mathcal{N}$ starting with the preflow $f_l'$ formed by increasing $f_l$ on arcs $(s,v)$ to saturate them and decreasing $f_l$ on arcs $(v,t)$ to meet the capacity constraints for $v\in U$. As an initial valid labeling, use $d(v)$$=$$\min\{d_{f_l'}(v,t),d_{f_l'}(v,s)$$+(|U|$$-$$2)\}$. Find the minimal and maximal minimum cuts $(C,\overline{C})$ and $(C',\overline{C}')$ for $\tilde{\alpha}$, respectively.
\STATE If $\overline{C}'$$=\{t\}$, set $\boldsymbol{\tau}_{A_u\setminus A_l}^*$$\leftarrow$$F(A_u)$$-$$F(A_l)$. Otherwise, run {\em Slice}$(\mathcal{N}(C'),\tilde{\alpha},\alpha_u,\tilde{f},f_u,C,A_u)$. And if $C\neq\{s\}$, then run {\em Slice}$(\mathcal{N}(C),\alpha_l,\tilde{\alpha},f_l,\tilde{f},A_l,\overline{C}')$.
\end{algorithmic}
\caption{Parametric preflow algorithm for the computation of $\prox_{\lambda\Omega_{F,p}}(\boldsymbol{z})$.}
\label{alg:prox_paraflow}
\end{algorithm}

As mentioned above, if the penalty is network representable, then problem \eqref{eq:para_prob} is solved as a parametric maxflow problem on network $\tilde{\mathcal{N}}$, where capacities $c_\alpha(s,v)$ for $v\in V$ are $c_\alpha(s,v)=(\phi_i(\alpha)+\text{const.})$ and the others are constants for $\alpha$ (note that
$\phi_i(\alpha)\geq 0$ for $\alpha\in J$). Since $\psi_{i}$ is convex, those capacities satisfy the conditions of {\em the monotone source-sink class of problems}, i.e.,
\begin{enumerate}
\item $c(s,v)$ is a non-decreasing function of $\alpha$ for all $v\in U$,
\item $c(v,t)$ is a non-increasing function of $\alpha$ for all $v\in U$, and
\item $c(u,v)$ is constant for all $u,v\in U\setminus\{s,t\}$.
\end{enumerate}
Therefore, for a given on-line sequence of parameter values $\alpha_{1}<\cdots<\alpha_{k}$, there exists a parametric maxflow algorithm that computes minimum cuts $(A_1,\overline{A}_1),\cdots,$ $(A_k,\overline{A}_k)$ on the network such that $A_1\subseteq\cdots\subseteq A_k$, and runs at the cost of only a constant factor in the worst-case time bound of a single maxflow computation.

If parametric capacities in the monotone source-sink class of problems are linear for $\alpha$, all {\em breakpoints}, i.e., a value of parameter $\alpha$ at which the capacity for the corresponding cut changes, can also be found at the cost of a constant factor in the worst-case time bound of a single maxflow computation using the GGT-type algorithms. However, this is generally not true for non-linear capacities because we must solve nonlinear equations to identify such a parameter value \cite{GMQT12}. Although this is the case for our situation in general, we can find such a value in closed-form for the important cases $p=2,+\infty$ due to its specific form of the problem.
\begin{lemma}
\label{lem:param}
For network $\tilde{\mathcal{N}}$ corresponding to a graph-representative penalty, the value of $\alpha$ such that
\begin{equation*}
{\textstyle \sum_{v\in V}}c_\alpha(s,v) = {\textstyle \sum_{v\in U\setminus \{t\}}}c(v,t)-{\textstyle \sum_{v\in U\setminus V}}c(s,v)
\end{equation*}
is found in close form for $p=2,+\infty$.
\end{lemma}
The concrete derivations of these closed-forms are described in Appendix~\ref{app:detail}. For the other cases, we can at least apply some line search for finding such value of $\alpha$ due to the monotonicity of $\phi_i$. Thus, we can adapt the procedure of the GGT-type algorithms to find the chain of solutions $A_1\subseteq\cdots\subseteq A_k$, which results in giving an optimal solution to problem~\eqref{eq:prox_lp}, as shown in Algorithm~\ref{alg:prox_paraflow} (a brief review on the preflow-push algorithm used in Algorithm~\ref{alg:prox_paraflow} is given in Appendix~\ref{app:preflow}).
\begin{theorem}
\label{th:cost_paraflow}
Algorithm~\ref{alg:prox_paraflow} is correct, and runs at the cost of a constant factor in the worst-case time bound of a single maxflow computation. For example, it runs in $O(dm\log(d^2/m))$ with dynamic trees.
\end{theorem}
That is, although the preflow algorithm is applied several times, the total runtime of Algorithm~\ref{alg:prox_paraflow} is equivalent to that of a single application of the preflow algorithm to the original network.

The interval $(\alpha_0,\alpha_{k+1})$ is chosen such that it covers all possible breakpoints $\alpha_1,\ldots,\alpha_k$. In other words, it suffices to select a sufficiently small $\alpha_0$ so that for each vertex $v$ such that $(s,v)$ is of nonconstant capacity, $c_{\alpha_0}(s,v)+\sum_{u\in U\setminus\{s,t\}}$$c(u,v)$$<$$c(v,t)$, which is given as
\begin{equation}
\label{eq:alpha0}
\alpha_0\leftarrow \psi_i'({\textstyle\min_{v\in V}}\{c(v,t)-{\textstyle\sum_{u\in V\setminus\{s,t\}}}c(u,v)\})-1.
\end{equation}
Similarly, it suffices to select $\alpha_{k+1}$ sufficiently large so that for each vertex $v$ such that $(s,v)$ is of nonconstant capacity, $c(v,t)+$$\sum_{u\in U\setminus\{s,t\}}c(v,u)\,$$<\,$$c_{\alpha_{k+1}}$$(s,v)$, which is obtained as $\alpha_{k+1}\leftarrow 0$.

By following the above results, any GGT-type algorithms can be adapted to solve the problem~\eqref{eq:para_prob}. Algorithm~\ref{alg:prox_paraflow} shows an adaptation of the simplified version \cite{BDG+07} of the original GGT algorithm.

\section{Related Work}
\label{sec:related}

Learning with structured sparsity-inducing regularization has been actively discussed in machine learning for a decade. Typical instances include (generalized) fused Lasso \cite{TSR+05,TT11} and group Lasso \cite{YL06,HZM11,BJMO12}. Generalized fused Lasso is closely related to the so-called total variation, which has often been discussed in computer vision \cite{ROF92}. Recently, group penalties have been applied to more complex groups, such as hierarchical penalty \cite{ZRY09,JMOB11} and path penalty \cite{MY13}. The total variation regularization is known to be solvable with an efficient parametric maxflow algorithm \cite{CD09,GY09}. In addition, the proximal problem for $l_1/l_\infty$-group penalty is calculated via parametric maxflow optimization \cite{MJOB11}. 
The proposed optimization formulation includes these formulations as special cases. Bach~(2010) \cite{Bac10} and Bach~\&~Obozinski~(2012) \cite{OB12} revealed that many of the existing structured penalties are obtained as convex relaxations of submodular functions, and those proximal problems are formulated as separable convex minimization.


The sufficient condition in Section~\ref{sec:network} is closely related to the class of energy minimization problems solvable by {\em graph-cut algorithm} \cite{BVZ01,KZ04}. Energy minimization is a formulation of the maximum a posteriori (MAP) estimation on MRFs (see, for example, \cite{WJ08}). Similar results are found in the context of realization of a submodular function as a cut function in combinatorial optimization \cite{BM85,FP01}.

Algorithm~\ref{alg:prox_paraflow} is a divide-and-conquer implementation of the preflow algorithm proposed by Gallo~\&~Tarjan~(1988) \cite{GT88}. Bach~(2010) \cite{Bac10} and Bach~\&~Obozinski~(2012) \cite{OB12} have mentioned an application of a divide-and-conquer approach to separable convex minimization proposed by Groenevelt~(1991)~\cite{Gro91} for proximal problem~\eqref{eq:prox_lp}, which takes $O(d)$ times of the cost for submodular minimization. Algorithm~\ref{alg:prox_paraflow} takes the cost for only a single run of the preflow algorithm by adapting Gallo~et~al.~(1989)~\cite{GGT89}'s algorithm to the current problem.

\section{Runtime Comparisons}
\label{sec:result}

\begin{figure}[t]
\centering
\subfigure[overlapping group penalty from $F(A)=\sum_g\min\{|A\cap g|,1\}(p=2)$]{
\includegraphics[width=.35\linewidth]{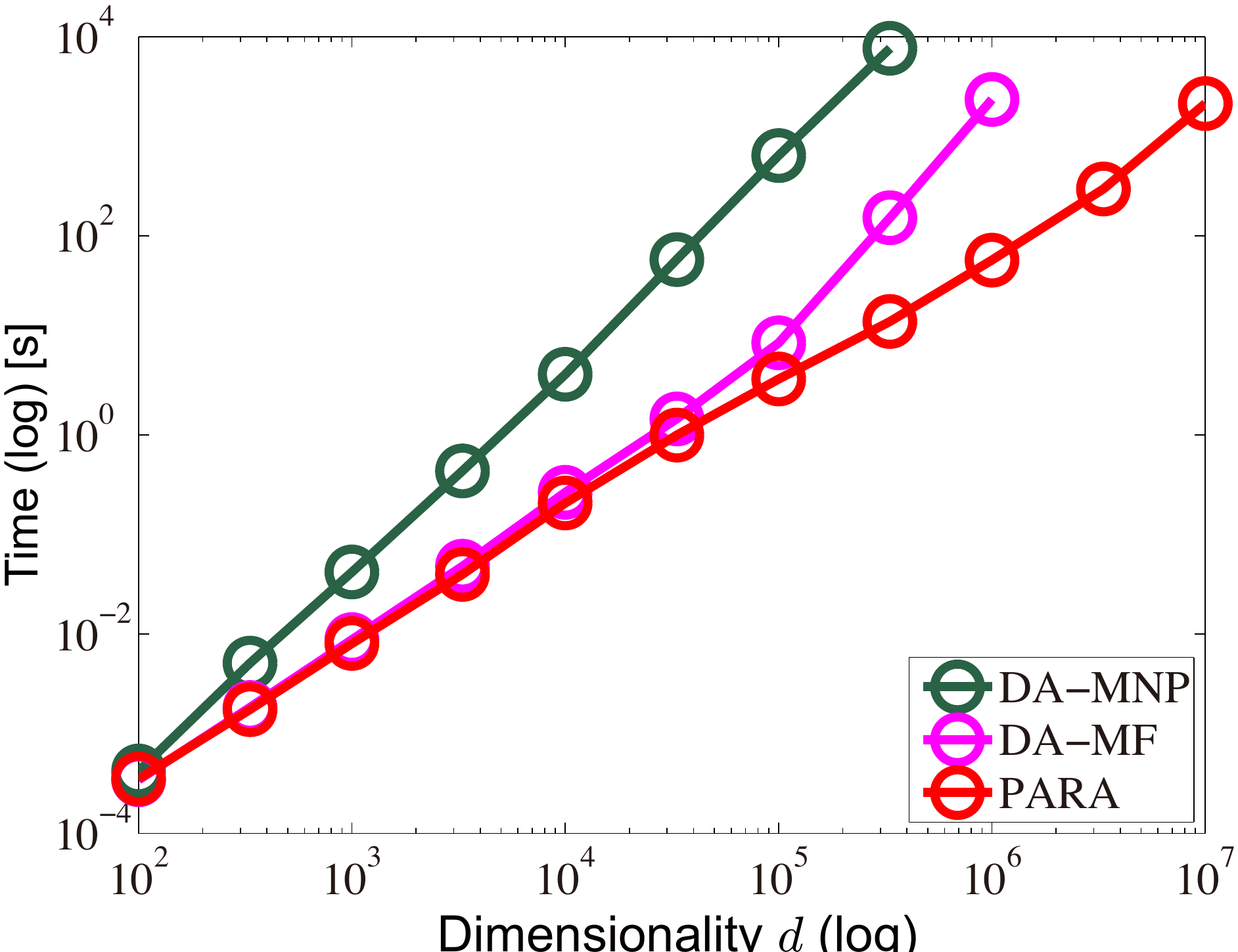}}
\hspace*{10mm}
\subfigure[overlapping $\ell_1/\ell_\infty$-group penalty]{
\includegraphics[width=.35\linewidth]{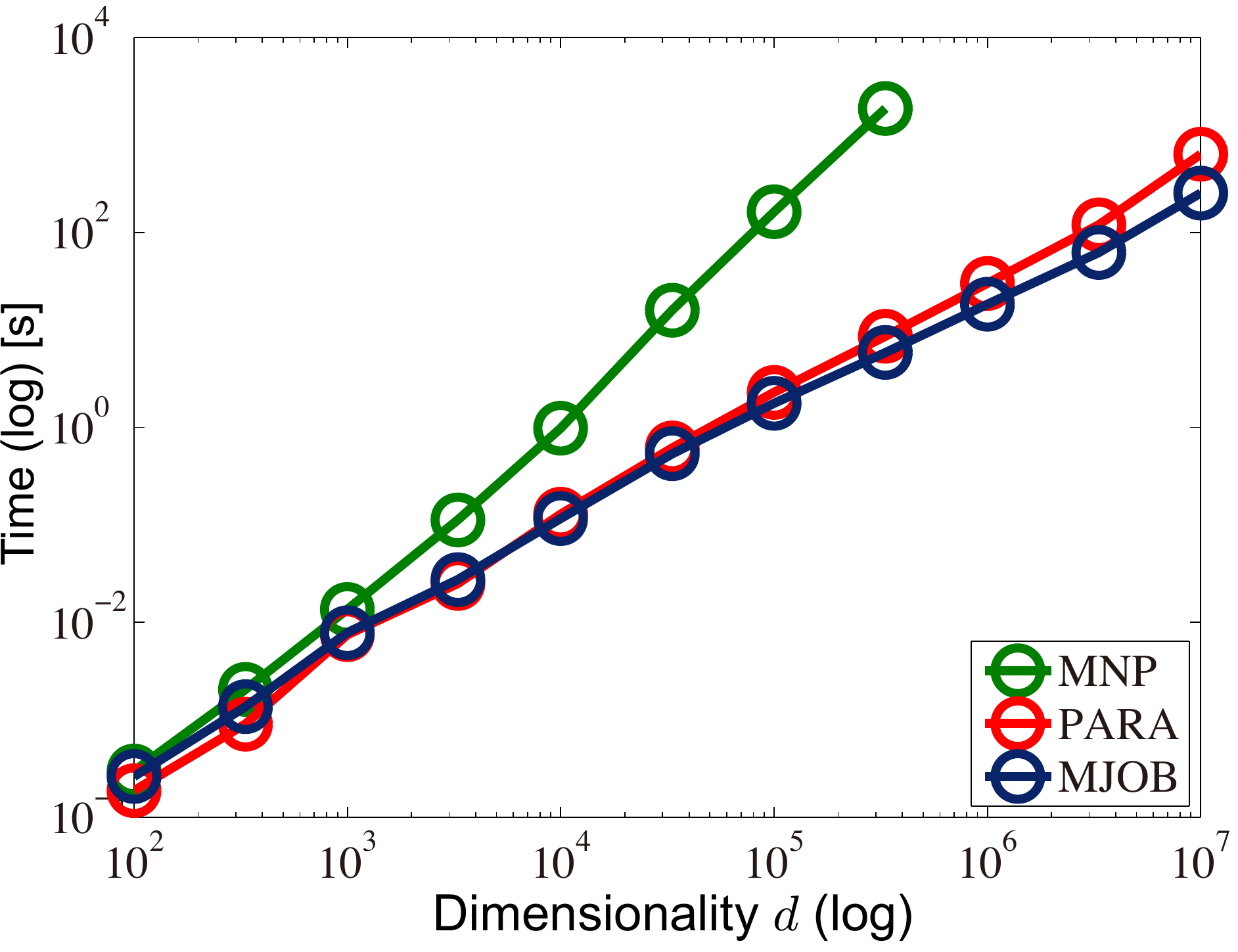}}
\vspace*{5mm}
\subfigure[fused penalty]{
\includegraphics[width=.35\linewidth]{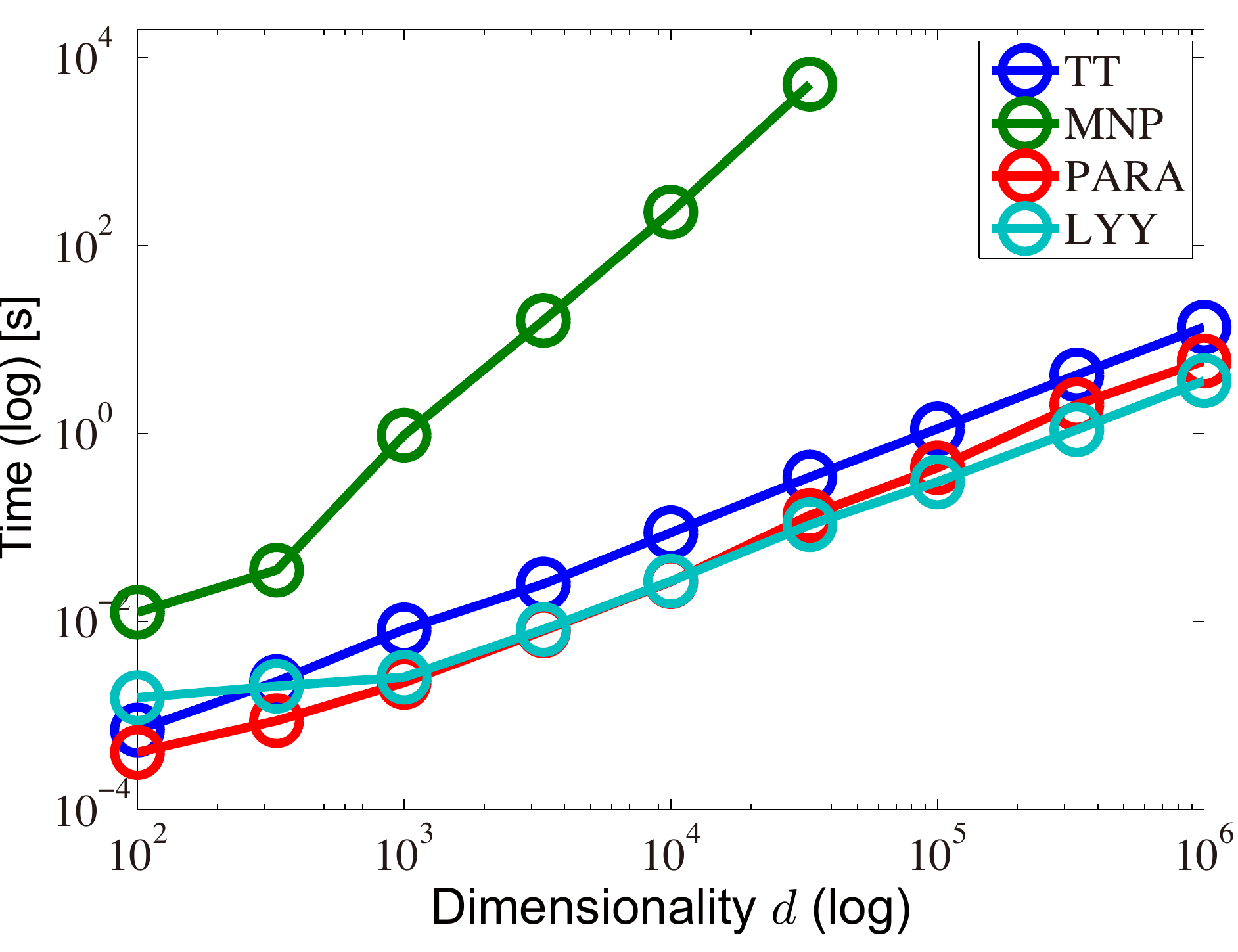}}
\hspace*{10mm}
\subfigure[generalized fused penalty]{
\includegraphics[width=.35\linewidth]{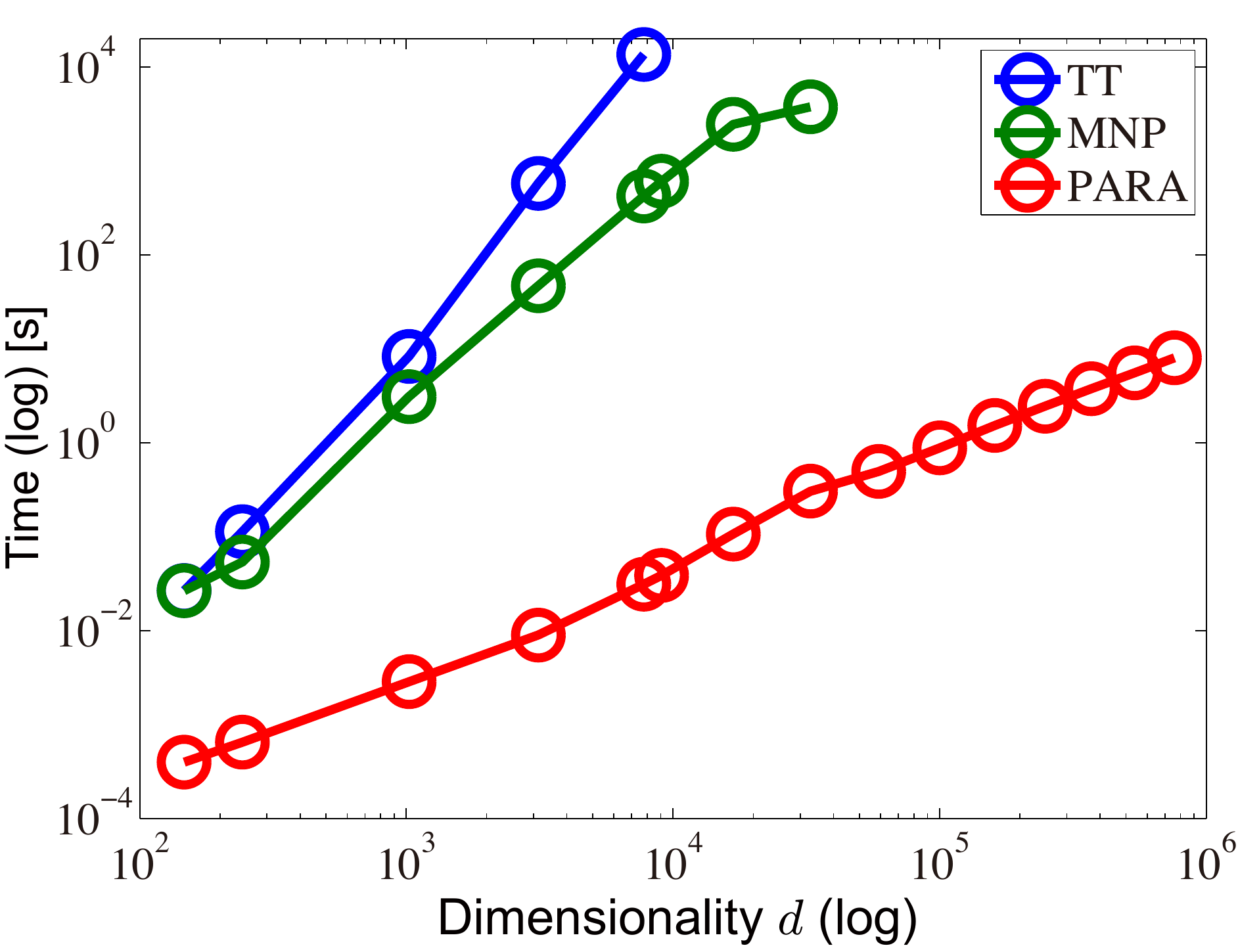}}\vspace*{-5mm}
\caption{Runtime comparisons for calculating $\prox_{\Omega_{F,p}}$.}
\label{fig:runtime}
\end{figure}

Here, we show empirical runtime comparisons of our algorithm with some existing ones based on different principles to see the scalabilities of the algorithms. 
The experiments were run on a 2.6 GHz 64-bit workstation using C++.\@ We applied our algorithm (we refer it as `PARA') to the proximal problems for the penalty from $F(A)=\sum_{g}\min\{|A\cap g|,1\}$ ($p=2,\infty$) (as a typical example of penalties described in 3.1.1) and the (generalized) fused penalty (as one described in 3.1.2). We compared ours with the following algorithms; the decomposition algorithm described in \cite{OB12,Bac13} with the minimum-norm-point (MNP) algorithm / the maxflow algorithm (`DA-MNP'/'DA-MF') and the algorithm by \cite{MJOB11} (`MJOB')\footnote{We used the code modified from the one available at http://spams-devel.gforge.inria.fr/} for the penalties from $F(A)=\sum_{g}\min\{|A\cap g|,1\}$ (MJOB is applicable only for $p$$=$$\infty$), and the MNP algorithm (`MNP'), the algorithm by Tibshirani~\&~Taylor~(2011)~\cite{TT11} (`TT') and the one by Liu~et~al.~(2010)~\cite{LYY10} (`LYY')\footnote{We used the code available at http://www.yelab.net/software/SLEP/} for the (generalized) fused penalty (LYY is applicable only to the 1d fused case). 

We generated data as follows. First, we generated a random vector $\mathbf{z}\in\mathbb{R}^d$ from the uniform distribution in $[-1,1]^d$. For generalized fused penalty, we randomly generated a directed network over nodes corresponding to $V$ using GENRMF from DIMACS Challenge.\footnote{The first DIMACS Int'l Algorithm Implementation Challenge (http://dimacs.rutgers.edu/Challenges/).} 
And for generating overlapping groups, we randomly generated $d/20$--$d/10$ groups of size $30$--$100$. 
The graphs in Figure~\ref{fig:runtime} show the empirical runtimes (in logarithm scale) for the algorithms. The plotted points are the averaged values over 10 randomly generated datasets.

\section{Conclusions}
\label{sec:concl}

In this paper, we provided a comprehensive class of structured penalties for which the proximal problem can be solved via an efficiently-solvable class of parametric maxflow optimization. Then, we showed that the parametric maxflow algorithm by Gallo~et~al.~(1989)~\cite{GGT89} and its variants, which runs at the cost of a constant factor in the worst-case time bound of the corresponding maxflow optimization, is applicable to solve this problem. The runtime of the proposed algorithm was empirically compared to those of the state-of-the-art ones.

Several avenues would be worth investigating: First, our formulation does not include the type of sparsity by the latent group penalties, such as \cite{JOV09}. As mentioned in \cite{MY13}, the proximal problem for the penalties of Jacob~et~al.~(2009)~\cite{JOV09} and its generalization can be solved as a minimum-cost flow problem, which is known to be calculated as a parametric maxflow problem if the costs are quadratic and only on edges connected to source/sink \cite{Hoc07}. It would be important to consider an unified framework connecting the current and such problems in the future work. Also, it would be interesting to address a new structured penalty satisfying the developed condition for some specific application.





\appendix

\section{Review of the Preflow-Push Algorithm}
\label{app:preflow}

The preflow algorithm computes a maximum flow in a directed network $\mathcal{N}$ \cite{GT88}. We first define terminology to describe the algorithm. A {\em preflow} $f$ on $\mathcal{N}$ is a real-valued function on vertex pairs satisfying the capacity constant, the antisymmetry constraint, and the following relaxation of the conservation constraint
\begin{equation}
\label{eq:nonnegativity}
\sum_{v_1\in U}f(v_1,v_2)\geq 0 ~\text{ for all }~ v_2\in V\setminus\{s,t\}.
\end{equation}
For a given preflow, we define the {\em excess} $e(v)$ of a vertex $v$ to be $\sum_{u\in U}f(u,v)$ if $v\neq s$, or infinity if $v=s$. We call a vertex $v\neq\{s,t\}$ {\em active} if $e(v)>0$. A preflow is a flow if and only if Eq.~\eqref{eq:nonnegativity} holds with equality for all $v\neq\{s,t\}$, i.e., $e(v)=0$ for all $v\neq\{s,t\}$. A vertex pair $(v,u)$ is a {\em residual arc} for $f$ if $(v,u)<c(v,u)$. A path of residual arcs is a {\em residual path}. A {\em valid labeling} $d$ for a preflow $f$ is a function from the vertices to the nonnegative integers and infinity, such that $d(t)=0$, $d(s)=n$, and $d(v)\leq d(u)+1$ for every residual arc $(v,u)$. The {\em residual distance} $d_f(v,u)$ from $v$ to $u$ is the minimum number of arcs on a residual path from $v$ to $u$, or infinity if there is no such a path.

To implement the preflow algorithm, we use the {\em incidence list} $I(v)$ for each vertex $v$. The elements of $I(v)$ are the unordered pairs $\{v,u\}$ such that $(v,u)\in E$ or $(u,v)\in E$. The algorithm consists of repeating the following procedure until no active vertices exist. Select any active vertex $v_1$. Let $(v_1,v_2)$ be the current edge of $v_1$. Then, apply the appropriate one of the following three cases.
\begin{description}
\item[{\it Push:}] If $d(v_1)>d(v_2)$ and $f(v_1,v_2)<c(v_1,v_2)$, send $\delta=\min\{e(v_1),c(v_1,v_2)-f(v_1,v_2)\}$ units of flow from $v_1$ to $v_2$, by increasing $f(v_1,v_2)$ and $e(v_2)$ by $\delta$, and by decreasing $f(v_1,v_2)$ and $e(v_1)$ by $\delta$.
\item[{\it Get Next Edge:}] If $d(v_1)\leq d(v_2)$ or $f(v_1,v_2)=c(v_1,v_2)$, and $(v_1,v_2)$ is not the last edge in $I(v_1)$, replace $(v_1$$,$$v_2)\,$as the current edge of $v_1$ with the next in$\,I(v_1)$.
\item[{\it Relabel:}] If $d(v_1)\leq d(v_2)$ or $f(v_1,v_2)=c(v_1,v_1)$, and $(v_1,v_2)$ is the last edge in $I(v_1)$, replace $d(v_1)$ by $\min\{(v_1,v_2)\in I(v_1),f(v_1,v_2)<c(v_1,v_2)\}$$+1$ and make the first edge in $I(v_1)$ the current edge of $v_1$.
\end{description}
When the algorithm terminates, $f$ is a maximum flow. A minimum cut can be computed, after replacing $d(v)$ by $\min\{d_f(v,s)+n,d_f(v,t)\}$ for each $v\in V$, as $(A,\overline{A})$ such that $A=\{v|d(v)\geq n\}$, where the sink side $\overline{A}$ is of minimum. The worst-case total time is $O(dm\log(d^2/m))$ if we use dynamic trees for the selection of active vertices.

\section{Details of Algorithm~\ref{alg:prox_paraflow}}
\label{app:detail}

In this appendix, we describe the details of Algorithm~\ref{alg:prox_paraflow} for solving the proximal problem~\eqref{eq:prox_lp}. Especially, we give the closed-form solutions for finding $\alpha$ described in Lemma~\ref{lem:param} for $p = 2,\infty$ (i.e., $r=1,2$), which is the key to make the complexity of Algorithm~\ref{alg:prox_paraflow} equivalent to the original GGT-type algorithm.

First, from the definition (see, Eq.~\eqref{eq:sepa_prob}), function $\psi_i(\tau_i)$ is represented as
\begin{equation*}
\psi_i(\tau_i) =
\begin{cases}
\frac{1}{2}\lambda^2\tau_i^{2/r}-\lambda\tau_i^{1/r}|z_i| & (0\leq \tau_i\leq \left(|z_i|/\lambda\right)^r) \\
-\frac{1}{2}z_i^2 & (\left(|z_i|/\lambda\right)^r < \tau_i).
\end{cases}
\end{equation*}
Note that this function is non-increasing for $\tau_i$ (for $\tau_i$ such that $0\leq \tau_i\leq \left(|z_i|/\lambda\right)^r$, it is monotone). The derivative is given by
\begin{equation}
\label{eq:psi_dev}
\psi_i'(\tau_i) = \begin{cases}_
\frac{\lambda^2\tau_i^{1/r}-\lambda|z_i|}{r\tau_i^{1-1/r}} & (0\leq \tau_i\leq \left(|z_i|/\lambda\right)^r) \\
0 & (\left(|z_i|/\lambda\right)^r < \tau_i).
\end{cases}
\end{equation}
This derivative is a non-decreasing function for $\tau_i$ (for $\tau_i$ such that $0\leq \tau_i\leq \left(|z_i|/\lambda\right)^r$, it is monotone). Hence, $\psi_i'$ has an inverse function for $0\leq \tau_i\leq \left(|z_i|/\lambda\right)^r$.

To give an closed-form solution for $\alpha$ as in Eq.~\eqref{eq:A2tau} and in Lemma~\ref{lem:param}, it is sufficient to describe  how we can find $\tilde{\alpha}$ satisfies for $S\subseteq V$
\begin{equation*}
{\textstyle \sum_{i\in S}} \phi_i(\tilde{\alpha}) = \tilde{c},
\end{equation*}
where $\tilde{c}$ is some constant, which is stated in following parts for $p=2,\infty$, respectively.

\paragraph{Case for $p=2$ ($r=2$)}
By substituting $r=2$ into Eq.~\eqref{eq:psi_dev}, we have for $0\leq \tau_i\leq (|z_i|/\lambda)^2$
\begin{equation*}
\psi_i'(\tau_i) = \frac{\lambda}{2}\left(\lambda-|z_i|/\tau_i^{1/2}\right).
\end{equation*}
Therefore, $\tilde{\tau}_i$ and $\tilde{\tau}_j$ such that $\psi_i'(\tilde{\tau}_i)=\psi_j'(\tilde{\tau}_j)$ satisfy
\begin{equation*}
|z_i|^2 \tilde{\tau}_j = |z_j|^2 \tilde{\tau}_i.
\end{equation*}
This means that, if $\tilde{\alpha}$ satisfies $\sum_{i\in S}\phi_i(\tilde{\alpha})=\tilde{c}$, then we have
\begin{equation*}
\phi_i(\tilde{\alpha}) = \frac{|z_i|^2}{\sum_{j\in S}|z_j|^2}\,\tilde{c}.
\end{equation*}
Thus, we can calculate such $\tilde{\alpha}$ as
\begin{equation*}
\tilde{\alpha} = \psi_i'\Bigl(|z_i|^2\tilde{c}/{\textstyle\sum_{j\in S}}|z_j|^2\Bigr).
\end{equation*}

\paragraph{Case for $p=+\infty$ ($r=1$)}
By substituting $r=1$ into Eq.~\eqref{eq:psi_dev}, we have for $0\leq \tau_i\leq |z_i|/\lambda$
\begin{equation*}
\psi_i'(\tau_i) = \lambda (\lambda\tau_i-|z_i|).
\end{equation*}
Thus, $\tilde{\tau}_i$ and $\tilde{\tau}_j$ such that $\psi_i'(\tilde{\tau}_i)=\psi_j'(\tilde{\tau}_j)$ satisfy
\begin{equation*}
|z_i|-|z_j| = \lambda(\tilde{\tau}_i-\tilde{\tau}_j).
\end{equation*}
This means that, if $\tilde{\alpha}$ satisfies $\sum_{i\in S}\phi_i(\tilde{\alpha})=\tilde{c}$, then we have
\begin{equation*}
\phi_i(\tilde{\alpha}) = \frac{\tilde{c}}{d} + \frac{|z_i|-\sum_{j\in S}|z_j|/|S|}{\lambda}.
\end{equation*}
Hence, we can calculate such $\tilde{\alpha}$ as
\begin{equation*}
\tilde{\alpha} = \psi_i'\Biggl(\frac{\tilde{c}}{d} + \frac{|z_i|-\sum_{j\in S}|z_j|/|S|}{\lambda}\Biggr).
\end{equation*}

\section{Proofs}
\label{app:proofs}

\subsection*{Theorem~\ref{th:sufficient}}

\paragraph{{\rm (i)}}
Let $\tilde{\mathcal{N}}$ be the constructed network (see Figure~\ref{fig:graph}-(a)) with the additional node $u$. Then, for each $A \subseteq V$, we have $\kappa_{\tilde{\mathcal{N}}}(\{s\} \cup A) = \boldsymbol{w}(A)$ and $\kappa_{\tilde{\mathcal{N}}}(\{s\} \cup A \cup \{u\}) = y$. Hence, the constructed network indeed represents $F(A) = \min\{\boldsymbol{w}(A), y\}$.

\paragraph{{\rm (ii)}}
Let $\tilde{\mathcal{N}} = (U = V \cup W \cup \{s, t\}, \tilde{E})$ be the constructed network (see Figures~\ref{fig:graph}-(c),(d)). We show that $F(A) = \min_{Y \subseteq W} \kappa_{\tilde{\mathcal{N}}}(\{s\} \cup A \cup Y) - \kappa_{\tilde{\mathcal{N}}}(\{s\}) + F(\emptyset)$ for every $A \subseteq V$. It is easy to confirm that, for each $A \subseteq V$, the set $\{\, w_B \in W \mid B \subseteq A \,\}$ attains the minimum of $\min_{Y \subseteq W}\kappa_{\tilde{\mathcal{N}}}(\{s\} \cup A \cup Y)$. When $A = \emptyset$, the minimum value is indeed $\kappa_{\tilde{\mathcal{N}}}(\{s\})$. Besides, when $\emptyset \neq A \subseteq V$, it increases by $\sum_{v \in V} \max\{0, F^{(1)}(v)\}$ and decreases by $\sum_{v \in V} \max\{0, -F^{(1)}(v)\}$ and $\sum_{A} \{\, -F^{(|A|)}(A) \mid A \subseteq V \text{ with } |A| \geq 2 \,\}$, which implies that $\min_{Y \subseteq W}\kappa_{\tilde{\mathcal{N}}}(\{s\} \cup A \cup Y) = \kappa_{\tilde{\mathcal{N}}}(\{s\}) + \sum_{A} \{\, F^{(|A|)}(A) \mid \emptyset \neq A \subseteq V \,\} = \kappa_{\tilde{\mathcal{N}}}(\{s\}) + F(A) - F(\emptyset)$.

\paragraph{{\rm (iii)}}
For a fixed set function $F$ satisfying Condition (iii), we construct a directed network $\tilde{\mathcal{N}}=(U = V \cup W \cup \{s, t\}, \tilde{E})$ with nonnegative capacity $c \colon \tilde{E} \to \mathbb{R}_+$ as follows. Then, $\tilde{\mathcal{N}}$ coincides with the network just before Step 4 in the construction procedure in Section~\ref{ssec:cond} (up to modular terms), and we have $F(A) = \min_{Y\subseteq W}\kappa_{\tilde{\mathcal{N}}}(\{s\}\cup A\cup Y)$ for every $A \subseteq V$. 

First, we define $W$ as the union of the following:
\begin{equation*}
\begin{split}
W_{2} &:= \{\, w_{A} \mid A \in \textstyle\binom V 2 \,\}, \\
W_{3}^{+} &:= \{\, w_{A} \mid A \in \textstyle\binom V 3 \text{ with } F^{(3)}(A)>0 \,\}, \\
W_{3}^{-} &:= \{\, w_{A} \mid A \in \textstyle\binom V 3 \text{ with } F^{(3)}(A)<0 \,\},
\end{split}
\end{equation*}
where each $w_{A}$ is an additional node adjacent to the nodes in $A$. Next, we define $\tilde{E}$ as the union of the following:
\begin{equation*}
\begin{split}
& E_{1}^{+} := V \times \{t\},~ E_{1}^{-} := \{s\} \times V,\\
& E_{2} := \{s\} \times W_2,~ E_{21}:=\{\, (w_A, v) \mid w_A\in W_2,~ v\in A \,\},\\
& E_{3}^{+} := W_3^+ \times \{t\},~ E_{13}:=\{\, (v, w_A) \mid w_A \in W_3^+,~ v\in A \,\},\\
& E_3^- := \{s\} \times W_3^-,~ E_{31}:=\{\, (w_A, v) \mid w_A\in W_3^-,~ v\in A \,\}.
\end{split}
\end{equation*}
Let us define a set function $H \colon 2^V \to \mathbb{R}$ as
\begin{equation*}
\begin{split}
H(A) := {\textstyle\sum_{B}}\{\, F^{(3)}(B) \mid A\subseteq B \subseteq V,~ w_B\in W_3^+ \,\}~~(A \subseteq V),
\end{split}
\end{equation*}
and the capacity function $c \colon E \to \mathbb{R}_+$ as, for each $e \in E$,
\begin{equation*}
\begin{split}
c(e) := \begin{cases}
\max\{0,F^{(1)}(\{v\})-H(\{v\})\} & (e=(v,t)\in E_1^+) \\
\max\{0,-F^{(1)}(\{v\})+H(\{v\})\} &(e=(s,v)\in E_1^-) \\
-F^{(2)}(A)-H(A) & (e=(s,w_A)\in E_2) \\
F^{(3)}(A) & (e=(w_A,t)\in E_3^+) \\
-F^{(3)}(A) & (e=(s,w_A)\in E_3^-) \\
+\infty & (e\in E_{21}\cup E_{13}\cup E_{31}).
\end{cases}
\end{split}
\end{equation*}
The nonnegativity of $c$ is guaranteed by the submodularity of $F$ as follows: for any $A = \{u, v\} \subseteq V$ with $|A| = 2$, we have
\begin{equation*}
\begin{split}
0 &\leq \min_{B}\{\, F(B \setminus \{u\}) + F(B \setminus \{v\}) - F(B) - F(B \setminus \{u, v\}) \mid A \subseteq B \subseteq V \,\}\\
&= \min_{B}\left\{\, -\sum_{B'}\{\, F^{(|B'|)}(B') \mid A \subseteq B' \subseteq B\,\} \ \Biggl| \ A \subseteq B \subseteq V \,\right\}\\
&= \min_{B}\left\{\, -F^{(2)}(A) - \sum_{B'}\left\{\, F^{(3)}(B') \ \Bigl| \ A \subseteq B' \in \binom B 3 \,\right\} \ \Biggl| \ A \subseteq B \subseteq V \,\right\}\\
&= -F^{(2)}(A) - \max_{\tilde{B} \subseteq V \setminus A}\sum_{v \in \tilde{B}} F^{(3)}(A \cup \{v\})\\
&= -F^{(2)}(A) - H(A).
\end{split}
\end{equation*}

We first check the value of $\min_{Y\subseteq W}\kappa_\mathcal{N}(Y\cup\{s\})$. If $Y\cap(W_2\cup W_3^-)\neq\emptyset$, then at least one edge in $E_{21}\cup E_{31}$ contributes to the cut capacity, which makes it $+\infty$. Otherwise (i.e., if $Y\subseteq W_3^+$), as no edge in $E$ is from $s$ to $W_3^+$, we have $\kappa_\mathcal{N}(Y\cup\{s\})\geq\kappa_\mathcal{N}(\{s\})$ for any $Y\subseteq W_3^+$, which means that $Y=\emptyset$ attains the minimum value. Without loss of generality, we assume $F^{(0)}(\emptyset) = F(\emptyset) = \kappa_\mathcal{N}(\{s\})$
(i.e., $C_F = 0$),
Then, it suffices to show that $F(A) = \min_{Y\subseteq W}\kappa_{\tilde{\mathcal{N}}}(A\cup Y\cup\{s\})$ for each nonempty $A\subseteq V$.

For any $B \subseteq V$ with $w_B \in W_2\cup W_3^-$, only the edge $(s, w_B)$ enters $w_B$, and the edges $(w_B, v)$ $(v \in B)$ with $c(w_B, v) = +\infty$ leave $w_B$. Therefore, if $B\subseteq A$, we have
\begin{equation*}
\kappa_\mathcal{N}(A\cup Y\cup\{s,w_B\})
= \kappa_\mathcal{N}(A\cup Y\cup\{s\})-c(s, w_B)
\leq\kappa_\mathcal{N}(A\cup Y\cup\{s\})
\end{equation*}
for every $Y\subseteq W\setminus\{w_B\}$. Moreover, for any $B\subseteq V$ with $w_B\in W_3^+$, only the edge $(w_B, t)$ leaves $w_B$, and the edges $(v, w_B)$ $(v \in B)$ with $c(v, w_B) = +\infty$ enter $w_B$. Thus, if $B \cap A \neq\emptyset$, we have $w_B \in Y$ and the edge $(w_B, t)$ contributes to the cut capacity. Thus, the minimum value is attained by $Y:=\{\, w_B \in W_2\cup W_3^- \mid B \subseteq A \,\}\cup\{\, w_B \in W_3^+ \mid B \cap A\neq\emptyset \,\}$, and we have
\begin{equation*}
\begin{split}
 & \kappa_{\tilde{\mathcal{N}}}(A\cup Y\cup \{s\})-\kappa_{\tilde{\mathcal{N}}}(\{s\}) \\
&= \sum_{v\in A}(c(v, t)-c(s, v))-\hspace*{-5mm}\sum_{w_B\in W_2\cup W_3^-\colon B\subseteq A}\hspace*{-7mm}c(s, w_B)+\hspace*{-4mm}\sum_{w_B\in W_3^+\colon B\cap A\neq\emptyset}\hspace*{-6mm}c(w_B, t)\\
&= \sum_{v\in A}(F^{(1)}(v)-H(v))+\hspace*{-4mm}\sum_{w_B\in W_2 \colon B \subseteq A}\hspace*{-4mm}(F^{(2)}(B)+H(B)) + \hspace*{-3mm}\sum_{w_B\in W_3^-\colon B\subseteq A}\hspace*{-3mm}F^{(3)}(B)+\hspace*{-4mm}\sum_{w_B\in W_3^+\colon B\cap A\neq\emptyset}\hspace*{-3mm}F^{(3)}(B)\\
&= \hspace*{-2mm}\sum_{B\colon \emptyset\neq B\subseteq A}\hspace*{-3mm}F^{(|B|)}(B) + \hspace{-4mm} \sum_{w_B\in W_3^+ \colon B\cap A\neq\emptyset\neq B\setminus A} \hspace{-6mm} F^{(3)}(B)~-~\sum_{v\in A}H(v)~+~\hspace*{-5mm}\sum_{w_B\in W_2\colon B\subseteq A}\hspace*{-5mm}H(B) \\
&= F(A)-F^{(0)}(\emptyset),
\end{split}
\end{equation*}
which means $\kappa_{\tilde{\mathcal{N}}}(A\cup Y\cup\{s\})=F(A)$. To see the last equality, it suffices to count the contribution of $F^{(3)}(B')$ to the second to the last line, which is easily seen to be totally zero, for each $B'\subseteq V$ with $w_{B'}\in W_3^+$.


\subsection*{Lemma~\ref{lem:non_dec}}

The first is shown in Lemma~2~and~3 in \cite{NK13} or Proposition~2.5 in \cite{Bac13}. The equivalence of optimal solutions to the two problems is obvious.

\subsection*{Corollary~\ref{cor:w_star}}

First, from Proposition~8.8 in \cite{Bac13}, we obtain a solution to problem~\eqref{eq:sepa_prob} as
\begin{equation}
t^*_i = \begin{cases}
\tau_i^* & \text{if}~~ (|z_i|/\lambda)^r>\tau_i^*, \\
\text{sign}(z_i)(|z_i|/\lambda)^r & \text{otherwise}.\vspace*{-.5mm}
\end{cases}
\end{equation}
Although the proposition assumes the strict convexity on separable functions, the above can be obtain since $(-\psi_i)'$ is monotone for $\tau_i$ s.t.\@ $(|z_i|/\lambda)^r>\tau_i^*$. Then, the corollary follows by solving analytically the minimization w.r.t.\@ $w$ in the definition of $\psi_i$.\vspace*{-1mm}

\subsection*{Lemma~\ref{lem:param}}
The statement of this lemma is shown by Appendix~\ref{app:detail}.

\subsection*{Theorem~\ref{th:cost_paraflow}}
The correctness follows the monotone source-sink property of the current network. The runtime follows the analysis in \cite{GGT89} from Lemma~\ref{lem:param}.

\bibliographystyle{plain}
\bibliography{proxflow_jmlr}

\end{document}